\documentclass{article} 
\usepackage{iclr2025_conference,times}

\usepackage{amsmath,amsfonts,bm}

\def\eqref#1{equation~\ref{#1}}

\def\1{\bm{1}}

\DeclareMathAlphabet{\mathsfit}{\encodingdefault}{\sfdefault}{m}{sl}
\SetMathAlphabet{\mathsfit}{bold}{\encodingdefault}{\sfdefault}{bx}{n}

\DeclareMathOperator*{\argmin}{arg\,min}

\usepackage[pagebackref,breaklinks,colorlinks]{hyperref}
\usepackage{hyperref}
\usepackage[accsupp]{axessibility}  
\usepackage{url}            
\usepackage{booktabs}       
\usepackage{amsfonts}       
\usepackage{nicefrac}       
\usepackage{xcolor}       
\usepackage{xkcdcolors}
\usepackage{epsfig}
\usepackage{graphicx}
\usepackage{placeins}
\usepackage{multirow}
\usepackage[skip=5pt]{caption}
\usepackage{rotating}
\usepackage{cancel}
\usepackage{setspace}
\usepackage{eso-pic}

\usepackage{bm}
\usepackage{bibunits} 

\usepackage{amssymb}
\usepackage{amsmath}
\usepackage{graphicx}
\usepackage{wrapfig,lipsum,booktabs}

\usepackage{epsfig}
\usepackage{tikz}
\usepackage{algpseudocode}
\usepackage{algorithm}
\usepackage{mathrsfs}

\usepackage{booktabs}       

\usepackage{thmtools,thm-restate}
\usepackage[capitalize,noabbrev]{cleveref}
\usepackage{listings}
\usepackage{lstautogobble}  %
\usepackage{color}          %
\usepackage{colortbl}
\usepackage{zi4}            %

\usepackage{subcaption}        
\usepackage{siunitx}               
\usepackage[export]{adjustbox}
\usepackage{makecell}
\usepackage{url}
\usepackage{tikz, pgfplots}
\usetikzlibrary{positioning}
\usepackage{capt-of}
\usepackage{diagbox}

\usepackage{orcidlink}
\usepackage{slashbox}
\usepackage{pifont}

\newcommand{\xb}{{\boldsymbol x}}

\newcommand{\wb}{{\boldsymbol w}}
\newcommand{\nb}{{\boldsymbol n}}

\newcommand{\fb}{{\boldsymbol f}}

\newcommand{\f}{{\boldsymbol f}}
\newcommand{\s}{{\boldsymbol s}}
\newcommand{\n}{{\boldsymbol n}}
\newcommand{\x}{{\boldsymbol x}}
\newcommand{\y}{{\boldsymbol y}}
\newcommand{\z}{{\boldsymbol z}}
\newcommand{\w}{{\boldsymbol w}}

\newcommand{\epsilonb}{{\boldsymbol \epsilon}}

\newcommand{\Ib}{{\boldsymbol I}}

\newcommand{\Ed}{{\mathbb E}}
\newcommand{\Rd}{{\mathbb R}}

\newcommand{\Ac}{{\mathcal A}}

\newcommand{\Nc}{{\mathcal N}}

\newcommand{\add}[1] {\textcolor{black}{#1}} 

\definecolor{trolleygrey}{rgb}{0.5, 0.5, 0.5}
\definecolor{BrickRed}{rgb}{0.6,0,0}
\definecolor{RoyalBlue}{rgb}{0,0,0.8}
\definecolor{Tdgreen}{rgb}{0,0.4,0.7}
\definecolor{pinegreen}{rgb}{0.0, 0.47, 0.44}
\definecolor{cornellred}{rgb}{0.7, 0.11, 0.11}
\definecolor{cadmiumgreen}{rgb}{0.0, 0.42, 0.24}
\definecolor{spirodiscoball}{rgb}{0.06, 0.75, 0.99}
\definecolor{mylightblue}{rgb}{0.85, 0.90, 0.94}
\definecolor{maroon}{cmyk}{0,0.87,0.68,0.32}

\renewcommand{\eqref}[1]{Eq.~(\ref{#1})}

\definecolor{C0}{rgb}{0.121569, 0.466667, 0.705882}
\definecolor{C1}{rgb}{1.000000, 0.498039, 0.054902}
\definecolor{C2}{rgb}{0.172549, 0.627451, 0.172549}
\definecolor{C3}{rgb}{0.839216, 0.152941, 0.156863}
\definecolor{C4}{rgb}{0.580392, 0.403922, 0.741176}
\definecolor{C5}{rgb}{0.549020, 0.337255, 0.294118}
\definecolor{C6}{rgb}{0.890196, 0.466667, 0.760784}
\definecolor{C7}{rgb}{0.498039, 0.498039, 0.498039}
\definecolor{C8}{rgb}{0.737255, 0.741176, 0.133333}
\definecolor{C9}{rgb}{0.090196, 0.745098, 0.811765}
\definecolor{trolleygrey}{rgb}{0.5, 0.5, 0.5}
\definecolor{darkpastelgreen}{rgb}{0.01, 0.75, 0.24}
\definecolor{darkpink}{rgb}{0.91, 0.33, 0.5}
\definecolor{alizarin}{rgb}{0.82, 0.1, 0.26}
\definecolor{americanrose}{rgb}{1.0, 0.01, 0.24}

\newcommand{\green}[1] {\textcolor{darkpastelgreen}{#1}}
\newcommand{\red}[1] {\textcolor{red}{#1}}
\newcommand{\cmark}{\green{\ding{51}}}%
\newcommand{\xmark}{\red{\ding{55}}}%

\usepackage{hyperref}
\usepackage{url}

\title{Amortized Posterior Sampling with \\ Diffusion Prior Distillation}

\author{Abbas Mammadov\thanks{Equal contribution}$^{\phantom{\ast},1,3}$, Hyungjin Chung$^{\ast,2,3}$, Jong Chul Ye$^{3}$ \\
$^1$Caltech, $^2$EverEx, $^3$KAIST \\
\texttt{\{abbas.mammadov, hj.chung, jong.ye\}@kaist.ac.kr}
}

\iclrfinalcopy 
\begin{document}

\maketitle

\begin{abstract}
    We propose Amortized Posterior Sampling (APS),
    a novel variational inference approach for efficient posterior sampling in inverse problems. Our method trains a conditional flow model to minimize the divergence between the variational distribution and the posterior distribution implicitly defined by the diffusion model. This results in a powerful, amortized sampler capable of generating diverse posterior samples with a single neural function evaluation, generalizing across various measurements. Unlike existing methods, our approach is unsupervised, requires no paired training data, and is applicable to both Euclidean and non-Euclidean domains. We demonstrate its effectiveness on a range of tasks, including image restoration, manifold signal reconstruction, and climate data imputation. APS significantly outperforms existing approaches in computational efficiency while maintaining competitive reconstruction quality, enabling real-time, high-quality solutions to inverse problems across diverse domains.    
\end{abstract}

\section{Introduction}
\label{sec:intro}

We consider the following inverse problem
\begin{align}
\label{eq:ip}
    \y = \Ac(\x) + \n ,\quad \y \in \Rd^m,\,\x\in\Rd^n,\,\Ac:\Rd^n \mapsto \Rd^m,\,\nb \sim \Nc(0,\sigma_y^2\Ib),
\end{align}
where the goal is to infer an unknown signal $\x$  from the degraded measurement $\y$ obtained through some forward operator $\Ac$, leveraging the information contained in the measurement and prior $p(\x)$. A powerful modern way to define the prior is through diffusion models~\citep{ho2020denoising,song2020score}, where we train parametrized model $\s_\theta$ to estimate the gradient of log prior $\nabla_\x \log p(\x)$. 

Solving inverse problems with the diffusion model can be achieved through posterior sampling with Bayesian inference. Arguably the standard way to achieve this is through modifying the reverse diffusion process of diffusion models~\citep{diffusion_survey}. This adjustment shifts the focus from sampling from the trained prior distribution $p_\theta(\x_0)$   to sampling from the posterior distribution  $p_\theta(\x_0|\y)$.  This transition is facilitated by employing iterative projections to the measurement subspace~\citep{kadkhodaie2020solving,song2020score,chung2022come,wang2023zeroshot}, guiding the samples through gradients pointing towards measurement consistency~\citep{chung2023diffusion,song2023pseudoinverseguided}. 

It should be noted that diffusion models learn the gradient of the prior and diffusion samplers~\citep{song2020denoising,lu2022dpm,song2020score} are methods that numerically solve the probability-flow ODE (PF-ODE) that defines the reverse diffusion sampling trajectory. Consequently, regardless of the specifics of the methods, standard diffusion model-based inverse problem solvers (DIS), even those that are considered {\em fast}, take at least a few tens of neural function evaluation (NFE), making them less effective for time-critical applications such as medical imaging and computational photography.

\begin{figure}[!htb]
\centering
\includegraphics[width=0.8\linewidth]{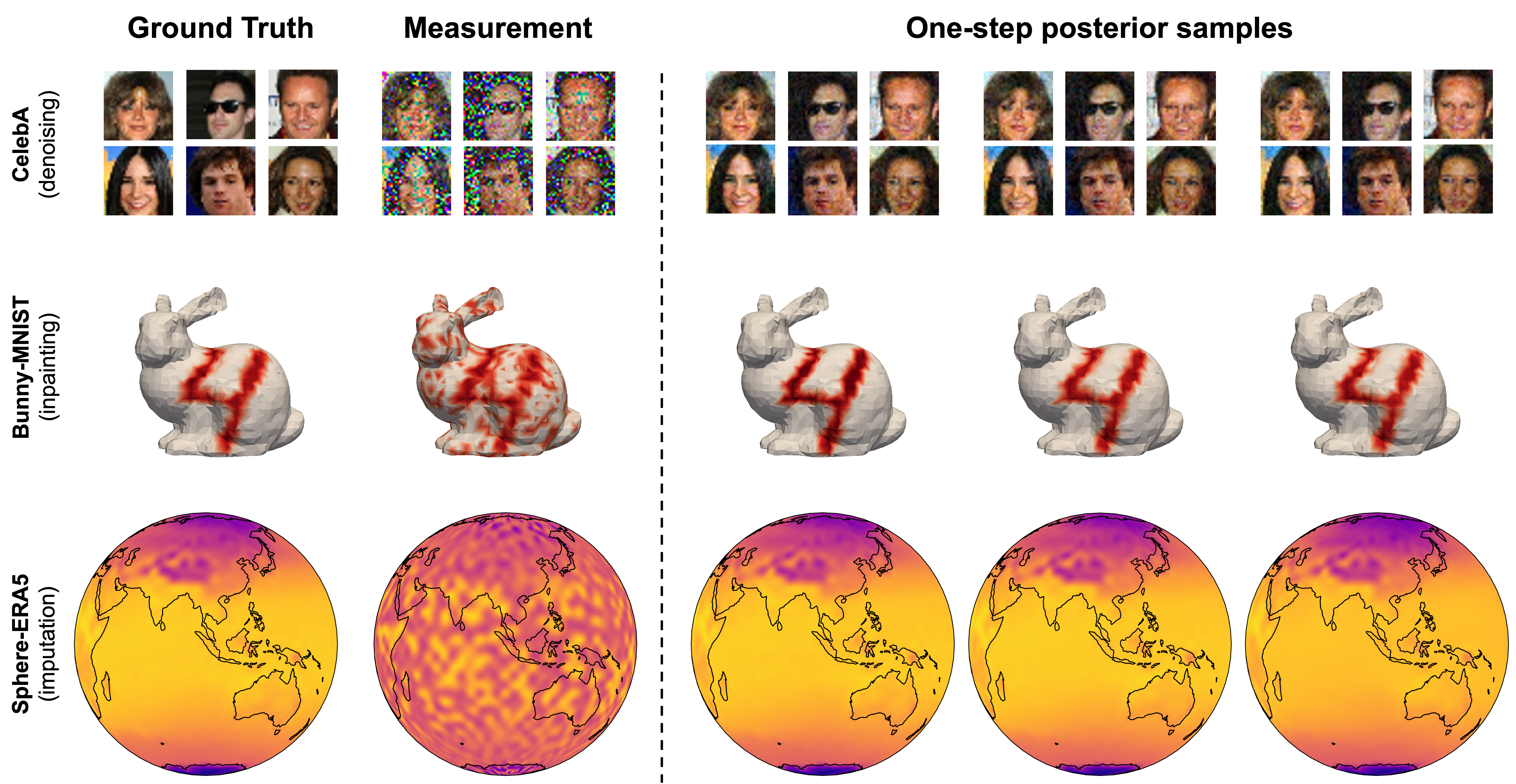}
\caption{Diverse inverse problem solving can be done with a \textbf{single} NFE, with the \textbf{same} network for all the different measurements. (row 1) Denoising on celebA, (row 2) inpainting MNIST on the bunny manifold, (row 3) imputation of ERA5 on the spherical manifold.}
\label{fig:main}
\vspace{-1em}
\end{figure}

Another class of methods~\citep{feng2023score,feng2024variational} introduces the use of variational inference (VI) to {\em train} a new proposal distribution $q_\phi^\y(\x)$ to {\em distill} the prior learned through the pre-trained diffusion model. The problem is defined as the following optimization problem
\begin{align}
\label{eq:vi}
    \min_\phi D_{KL}(q_\phi^\y(\x_0) || p_\theta(\x_0|\y)),
\end{align}
where the superscript $\y$ emphasizes that the proposal distribution is specific for a single measurement $\y$. For tractable optimization, $q$ is often taken to be a normalizing flow~\citep{rezende2015variational,dinh2016density} (NF) so that the computation of likelihood can be done instantly, and one can sample multiple reconstructions from the posterior samples by plugging in different noise values from the reference distribution. Posing the problem this way yields a method that can achieve posterior samples with just a single NFE. Nevertheless, it is still impractical as training a measurement-specific variational distribution takes hours of training. It is often unrealistic to train a whole new model from scratch every time when a new measurement is taken.

\begin{figure}[!htb]
\centering
\includegraphics[width=0.8\linewidth]{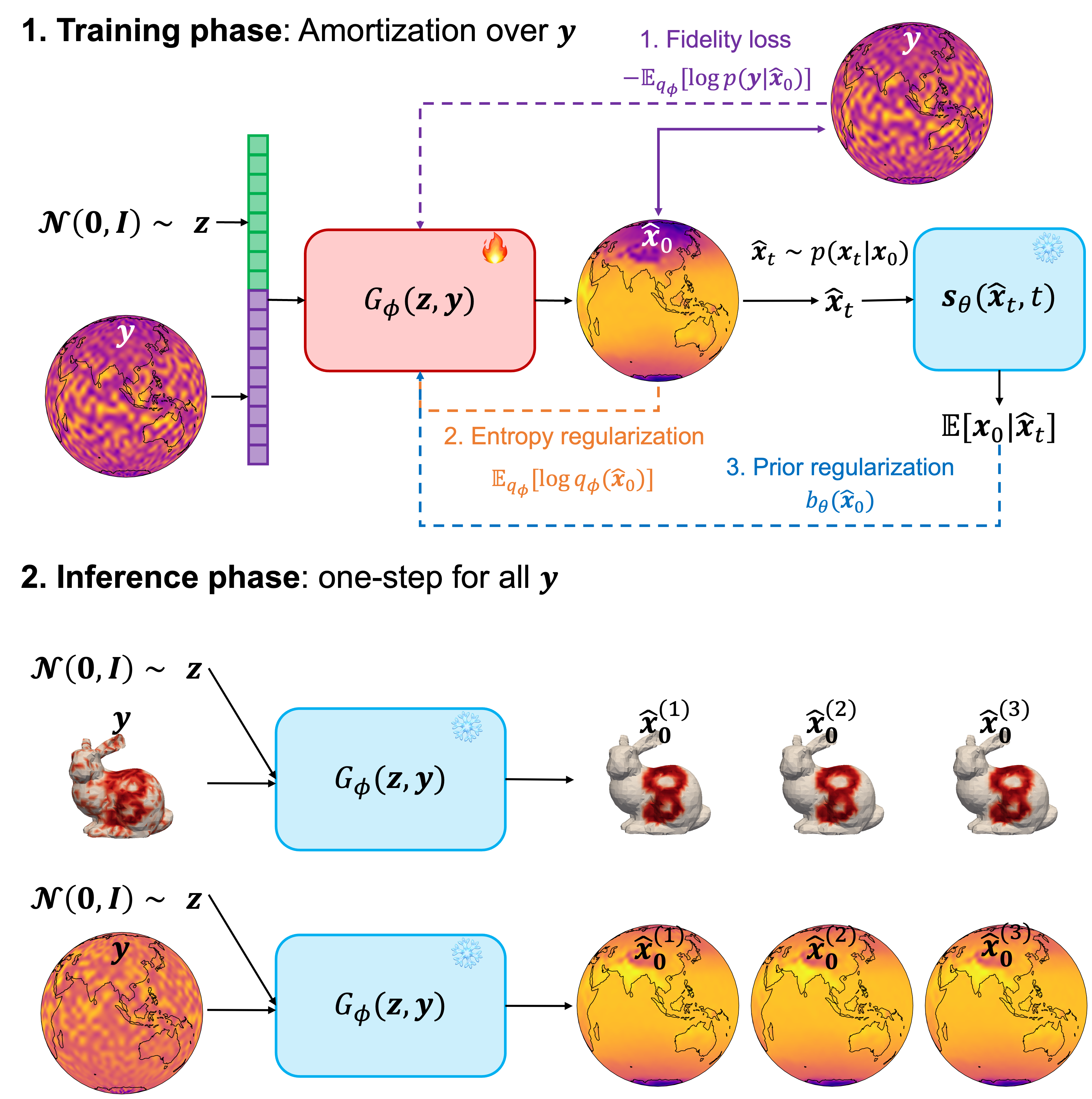}
\caption{Concept of the proposed method, APS. (a) Training can be performed in an unsupervised fashion with a dataset consisting of degraded measurements $\y$ to train a conditional normalizing flow $G_\phi$ with the diffusion prior $\s_\theta$. (b) Once trained, one can achieve multiple posterior samples by inputting different noise vectors $\z \sim \Nc(0, \Ib)$ concatenated with the condition $\y$ with a single NFE, generalizable across any measurement $\y$.}
\label{fig:concept}
\vspace{-1em}
\end{figure}

In this work, we take a step towards a {\em practical} VI-based posterior sampler by distilling a diffusion model prior. To this end, we propose a {\em conditional} normalizing flow $q_\phi(\x_0|\y)$ as our variational distribution and amortize the optimization problem in \eqref{eq:vi} over the conditions $\y$. By using a network that additionally takes in the condition $\y$ as the input, we can train a {\em single} model that generalizes across the whole dataset without the need for cumbersome re-training for specific measurements. (See Fig.~\ref{fig:concept} for the conceptual illustration of the proposed method, as well as representative results presented in Fig.~\ref{fig:main}.) Interestingly, we find that the speed of optimization is not hampered with such amortization, and the proposed method achieves comparable performance against the measurement-specific flow model~\citep{feng2024variational,feng2023score}. Furthermore, we extend the theory to consider inverse problems on the Riemannian manifold, showing that the proposed idea is generalizable even when the signal is not on the Euclidean manifold. In summary, our contributions and key takeaways are as follows

\begin{enumerate}
  \item We propose an amortized variational inference framework to enable 1-step posterior sampling constructed implicitly from the pre-trained diffusion prior $p_\theta(\x)$ for {\em any} measurement $\y$. 
  \item To the best of our knowledge, our method is the first diffusion prior distillation approach for solving inverse problems that are unsupervised (i.e. does not require any ground-truth data $\x$), as opposed to standard conditional NFs~\citep{lugmayr2020srflow} that required supervised paired data.
  \item Experimentally, we show that the proposed method easily scales to signals that lie on the standard Euclidean manifold (e.g. images) as well as signals that lie on the Riemannian manifold, achieving strong performance regardless of the representation.
\end{enumerate}

\section{Preliminaries}
\label{sec:prelinimaries}

\subsection{Score-based diffusion models}
\label{subsec:diffusion}

We adopt the standard framework for constructing a continuous diffusion process $\x(t)$, where $t \in [0, T]$ and $\x(t) \in \mathbb{R}^d$, as outlined by~\cite{song2020score}. Specifically, our goal is to initialize $\x(0)$ from a distribution $p_0(\x) = p_{\text{data}}$, and evolve $\x(t)$ towards a reference distribution $p_T$ at time $T$, which is easy to sample from.

The evolution of $\x(t)$ is governed by the Itô stochastic differential equation (SDE):
\begin{align}
\label{eq:forward-sde}
    d\x = \f(\x, t)dt + g(t)d\w,
\end{align}
where $\f: \mathbb{R}^d \times \mathbb{R} \rightarrow \mathbb{R}^d$ represents the drift function, and $g: \mathbb{R} \rightarrow \mathbb{R}^d$ denotes the diffusion coefficient. These coefficients are designed to drive $\x(t)$ towards a spherical Gaussian distribution as $t$ approaches $T$.
When the drift function $\f(\x, t)$ is affine, the perturbation kernel $p_{0t}(\x(t) | \x(0))$ is Gaussian, allowing for the parameters to be determined analytically. This facilitates data perturbation via $p_{0t}(\x(t) | \x(0))$ efficiently, without necessitating computations through a neural network.

Furthermore, corresponding to the forward SDE, there exists a reverse-time SDE:
\begin{align}
\label{eq:reverse-sde}
    d\xb &= [\fb(\xb, t) - g(t)^2 \nabla_\xb \log p_t(\xb)]dt + g(t) d\bar{\wb},
\end{align}
where $dt$ represents an infinitesimal negative time step, and $\bar{\wb}$ denotes the backward standard Brownian motion. While the trajectory of \eqref{eq:reverse-sde} is stochastic, there also exists a corresponding probability-flow ODE (PF-ODE) that recovers the same law $p_t(\x)$ as the time progresses~\citep{song2020score,song2020denoising}
\begin{align}
\label{eq:pf-ode}
    d\xb &= [\fb(\xb, t) - \frac{g(t)^2}{2} \nabla_\xb \log p_t(\xb)]dt.
\end{align}
This allows a deterministic mapping between the reference and the target distribution, and hence diffusion models can also be seen as a neural ODE~\citep{chen2018neural}.

A neural network can be trained to approximate the true score function $\nabla_\x \log p_t(\x)$ through score matching techniques, as demonstrated in previous works~\citep{song2019generative,song2020score}. This approximation, denoted $\s_{\theta}(\x, t) \approx \nabla_\x \log p_t(\x)$, is then utilized to numerically integrate the reverse-time SDE. To effectively train the score function, denoising score matching (DSM) is often employed~\citep{hyvarinen2005estimation}
\begin{align}
\label{eq:dsm}
    \theta^* = \argmin_\theta \Ed_{t \sim U(\varepsilon, 1), \x(t), \x(0)}\left[\|\s_\theta(\x(t), t) - \nabla_{\x_t}\log p_{0t}(\x(t)|\x(0))\|_2^2\right],
\end{align}
Interestingly, the posterior mean, or the so-called denoised estimate can be computed via Tweedie's formula~\citep{efron2011tweedie}. Specifically, for $p(\x_t|\x_0) = \Nc(\x_t; \alpha_t \x_0, \beta_t^2 \Ib)$, 
\begin{align}
\label{eq:tweedie}
    \hat\x_{0|t}^\theta := \Ed_{p(\x_0|\x_t)}[\x_0|\x_t] 
    = \frac{1}{\alpha_t}(\x_t + \beta_t^2 \nabla_{\x_t} \log p(\x_t)).
\end{align}

\subsection{Diffusion models for inverse problems (DIS)}
\label{subsec:dis}

Solving the reverse SDE in \eqref{eq:reverse-sde} or the PF-ODE in \eqref{eq:pf-ode} results in sampling from the prior distribution $p_\theta(\x_0)$, with the subscript emphasizing the time variable in the diffusion model context $\x_0 \equiv \x$. When solving an inverse problem as posed in \eqref{eq:ip}, our goal is to sample from the posterior $p_\theta(\x_0|\y) \propto p_\theta(\x)p(\y|\x_0)$. Using Bayes rule for a general timestep $t$ yields
\begin{align}
\label{eq:bayes}
    \nabla_{\x_t} \log p_\theta(\x_t|\y) = \nabla_{\x_t} \log p_\theta(\x_t) + \nabla_{\x_t} \log p(\y|\x_t).
\end{align}
While the former term can be replaced with a pre-trained diffusion model, the latter term is intractable and needs some form of approximation. Existing DIS~\citep{kawar2022denoising,chung2023diffusion,wang2023zeroshot} propose different approximations for $\nabla_{\x_t} \log p(\y|\x_t)$, which yields sampling from slightly different posteriors $\nabla_{\x_t} \log p_\theta(\x_t|\y)$.

Algorithmically, the posterior samplers are often implemented so that the original numerical solver for sampling from the prior distribution remains intact, while modifying the Tweedie denoised estimate at each time $\hat\x_{0|t}^\theta$ to satisfy the measurement condition given as \eqref{eq:ip}. From Tweedie's formula, we can see that this corresponds to approximating the conditional posterior mean $\Ed[\x_0|\x_t,\y]$ in the place of the unconditional counterpart $\Ed[\x_0|\x_t]$. The algorithms are inherently iterative, and the modern solvers~\citep{chung2023diffusion,wang2023zeroshot,zhu2023denoising} require at least 50 NFE to yield a high-quality sample. Moreover, as existing methods can be interpreted as approximating the reverse distribution $p(\x_0|\x_t)$ with a simplistic Gaussian distribution $q(\x_0|\x_t) = \Nc(\x_0; \hat\x_{0|t}^\theta, s_t^2\Ib)$~\citep{peng2024improving}, it typically yields a large approximation error, especially in the earlier steps of the reverse diffusion.

\begin{table}[t]
\centering
\setlength{\tabcolsep}{0.2em}
\resizebox{.75\textwidth}{!}{
\begin{tabular}{lc@{\hskip 10pt}c@{\hskip 30pt}c@{\hskip 10pt}c@{\hskip 10pt}c}
\toprule
{\textbf{Class}} & \multicolumn{2}{c}{\textbf{Score-based}} & \multicolumn{3}{c}{\textbf{Variational Inference}} \\
\cmidrule(lr){2-3}
\cmidrule(lr){4-6}
Methods & DIS & Noise2Score & RED-Diff  & Score prior & \textbf{APS (ours)} \\
\midrule
\thead{One-step inference} & \xmark & \cmark & \cmark & \cmark & \cmark \\
\thead{Tackles general\\inverse problems} & \cmark & \xmark & \xmark & \cmark & \cmark \\
\thead{Exact likelihood\\computation} & \xmark & \xmark & \xmark & \cmark & \cmark \\
\thead{Amortized across $\y$} & \xmark & \xmark & \xmark & \xmark & \cmark \\
\thead{Generalizable \\ across dataset} & \xmark & \xmark & \xmark & \xmark & \cmark \\
\thead{Blind sampling} & \xmark & \xmark & \xmark & \xmark & \cmark \\
\bottomrule
\end{tabular}
}
\caption{
Methods that leverage diffusion priors for solving inverse problems according to their class, and their characteristics.
}
\label{tab:taxonomy}
\vspace{-1em}
\end{table}

\section{Related works}
\label{sec:related_works}

\subsection{Variational inference in DIS}
\label{subsec:vi}

Standard DIS discussed in Sec.~\ref{subsec:dis} sample from the posterior distribution by following the reverse diffusion trajectory. Another less studied approach uses VI to use a new proposal distribution, where the problem is cast as an optimization problem in \eqref{eq:vi}.
RED-diff~\citep{mardani2023variational} places a unimodal Gaussian distribution as the proposal distribution $q_\phi^\y(\x)$, and the KL minimization is done in a coarse-to-fine manner, similar to standard DIS, starting from high noise level to low noise level. While motivated differently, RED-diff and standard DIS have similar downsides of requiring at least a few tens of NFEs, as well as placing a simplistic proposal distribution. Furthermore, one can achieve only a single sample per optimization.

Recently, Feng \emph{et al.}~\citep{feng2023score,feng2024variational} uses an NF model for the proposal distribution while solving the same VI problem. The optimization problem involves computing the diffusion prior log likelihood $\log p_\theta(\x)$. It was shown that it can be exactly computed by solving the PF-ODE~\citep{feng2023score,song2020score}, but numerically solving the PF-ODE per every optimization step is extremely computationally heavy, and hence does not scale well. To circumvent this issue, it was proposed to use a lower bound~\citep{feng2024variational,song2021maximum}. Once trained, the NF model can be given different noise inputs $\z \sim \Nc(0, \Ib)$ to generate diverse posterior samples with a single forward pass through the network. However, the training should be performed with respect to all the different measurements, not being able to generalize across the dataset. Our work follows along this path to overcome the current drawback and optimize a single model for entire measurement space with a similar cost as shown in Tab.~\ref{tab:taxonomy}.

\subsection{Distillation of the diffusion prior}
\label{subsec:distillation}

Our method involves distillation of the diffusion prior into a student deep neural network, in our case an NF model. Particularly, it involves evaluating the output of the model by checking the denoising loss gradients from the pre-trained diffusion model. This idea is closely related to variants of score distillation sampling (SDS)~\citep{poole2023dreamfusion,wang2024prolificdreamer}, where the gradient from the denoising loss is used to distill the diffusion prior by discarding the score Jacobian. Possibly a closely related work is Diff-instruct~\citep{luo2024diff}, where the authors propose to train a one-step generative model similar to GANs~\citep{goodfellow2014generative} by distillation of the diffusion prior with VI. By proposing an integral KL divergence (IKL) by considering KL minimization across multiple noise levels across the diffusion, it was shown that SDS-like gradients can be used to effectively train a new generative model. While having similarities, our method directly minimizes the KL divergence and does not require dropping the score Jacobian.

Orthogonal to the score distillation approaches, there have been recent efforts to train a student network to emulate the PF-ODE trajectory itself~\citep{song2023consistency,gu2023boot} with a single NFE, one of the most prominent directions being consistency distillation (CD)~\citep{song2023consistency}. While promising, the performance of CM is upper-bounded by the {\em teacher} PF-ODE. Thus, in order to leverage CD-type approaches for diffusion posterior sampling, one has to choose one of the approximations of DIS as its teacher model. In this regard, applying CD for diffusion inverse problem solving is inherently limited.

\section{Amortized Posterior Sampling (APS)}
\label{sec:main}

\subsection{Conditional NF for amortized score prior}
The goal is to use a variational distribution that is conditioned on $\y$, such that the resulting distilled conditional NF model $G_\phi$ generalizes to any condition $\y$. To this end, inspired from the choices of~\citep{sun2021deep,feng2023score} we modify the objective in \eqref{eq:vi} to
\begin{align}
    &\min_{\phi} D_{KL}(q_\phi(\x_0|\y) || p_\theta(\x_0|\y)) \\
    =&\min_\phi \int q_\phi(\x_0|\y)[-\log p(\y|\x_0) - \log p_\theta(\x_0) + \log q_\phi(\x_0|\y)]\\
    =&\min_\phi \Ed_{\z} \left[-\underbrace{\log p(\y|G_\phi(\z,\y))}_{\rm fidelity} - \underbrace{\log p_\theta(G_\phi(\z,\y))}_{\rm prior} + \underbrace{\log \pi(\z) - \log \left|\det \frac{dG_\phi(\z,\y)}{d\z}\right|}_{\rm induced\:entropy} \right],
\label{eq:conditional_score_prior}
\end{align}
where the second equality is the result of choosing a conditional NF as our proposal distribution, and now the expectation is over random noise $\z \sim \Nc(0, \Ib)$. Notice that our network takes in both a random noise $\z$ and the condition $\y$ as an input to the network.

Under the Gaussian measurement model in \eqref{eq:ip}, the fidelity loss reads
\begin{align}
\label{eq:fidelity}
    -\Ed_\z[\log p(\y|G_\phi(\z,\y))]
    &= -\Ed_\z\left[\frac{\|\y - \Ac(G_\phi(\z,\y))\|_2^2}{2\sigma_y^2}\right].
\end{align}
Moreover, the induced entropy can be easily computed as it is an NF
\begin{align}
\label{eq:entropy_flow}
    \Ed_{q_\phi(\x)}[\log q_\phi(\x)] = \Ed_{\z}\left[\log \pi(\z) - \log \left|\det \frac{dG_\phi(\z,\y)}{d\z}\right|\right]
\end{align}
where $\pi(z)$, in our case, is the reference Gaussian distribution $\Nc$. For simplicity, let us denote $\hat\x_0 := G_\phi(\z,\y)$.

 Computation of $\log p_\theta(\hat\x_0)$ is more involved: to exactly compute the value, we would have to solve the PF-ODE, which is compute-heavy~\citep{song2020score,feng2023score}. To circumvent this burden, we leverage the evidence lower bound (ELBO)~\citep{song2021maximum,feng2024variational} $b_\theta(\hat\x_0) \leq \log p_\theta(\hat\x_0)$:
\begin{align}
    b_\theta(\hat\x_0) = \Ed_{p(\hat\x_T|\hat\x_0)}\left[ \log \pi(\hat\x_T) \right] - \frac{1}{2} \int_0^T g(t)^2 h(t)\,dt
\end{align}
where
\begin{align}
    h(t) := \Ed_{p(\hat\x_t|\hat\x_0)}&\Bigg[
    \underbrace{\|\s_\theta(\hat\x_t) - \nabla_{\hat\x_t} \log p(\hat\x_t|\hat\x_0)\|_2^2}_{\rm DSM (\eqref{eq:dsm})} \notag
    \\ &- \| \nabla_{\hat\x_t} \log p(\hat\x_t|\hat\x_0) \|_2^2 - \frac{2}{g(t)^2} \nabla_{\hat\x_t} \cdot \fb(\hat\xb_t, t).
    \Bigg]
\end{align}
When we have $p(\x_t|\x_0) = \Nc(\x_t; \alpha_t \x_0, \beta_t^2\Ib)$ and a standard diffusion model with a linear SDE $\fb(\x_t,t) = f(t)\x_t$,
\begin{align}
    \| \nabla_{\x_t} \log p(\x_t|\x_0) \|_2^2 = \frac{1}{\beta(t)^2} \|\epsilonb\|_2^2,\quad
    \frac{2}{g(t)^2} \nabla_{\x_t} \cdot \fb(\xb_t, t) = \frac{2d\beta(t)}{g(t)^2},
\end{align}
where $d$ is the dimensionality of $\x_t$, and both terms are independent of $\phi$ and $\theta$. Intuitively, the DSM term evaluates the probability of $\x_0$ by measuring how easy it is to denoise the given $\x_0$. When the network easily denoises the given image, then it will assign a high probability. When not, a low probability is assigned.
We can now define an equivalent ELBO $b'_\theta(\x_0)$ in terms of optimization, which reads
\begin{align}
\label{eq:dsm_score_prior}
    b'_\theta(\x_0) = \Ed_{p_{0T}}[\log \pi(\x_T)] - \frac{1}{2}\int_0^T g(t)^2 \|\s_\theta(\x_t) - \nabla_{\x_t} \log p(\x_t|\x_0)\|_2^2\,dt
\end{align}
Plugging $b'_\theta(G_\phi(\z,\y))$ of \eqref{eq:dsm_score_prior} in the place of $\log p_\theta(G_\phi(\z,\y))$ in \eqref{eq:conditional_score_prior}, we can efficiently update $\phi$ by distilling the prior information contained in the diffusion model.

\subsection{Architecture}
It has been demonstrated in \citep{lugmayr2020srflow} that Conditional NFs are capable of learning distributions on the ambient space that are constrained on measurement. To achieve an architecture with invertible transformations, we extend RealNVP~\citep{dinh2016density} architecture to the conditional settings by borrowing insight from~\citep{sun2021deep}. In its plain form, RealNVP architecture mainly consists of Flow steps, each containing two Affine Coupling layers. In each affine coupling layer, input signal $\xb$ is split into two parts: $\xb_a$ which stays unchanged, and $\xb_b$ which is fed into the neural network. In order to invoke the condition, we simply concatenate the conditioning input $\y$ to the $\xb_b$ as these layers serve as the main and basic building blocks of entire invertible architecture. This seemingly simple integration led to very promising results in both Euclidean and non-Euclidean geometries as will be depicted in Section \ref{sec:exps}.

\subsection{Manifold}
Many real-world datasets, particularly in environmental science, naturally reside on non-Euclidean geometries, making inverse problems challenging. Our work extends conditional normalizing flows (CNFs) to distributions on non-Euclidean manifolds, enabling direct solving of inverse problems on these surfaces without additional rendering steps. We represent manifold data as point clouds of size $V \times C$, where $V$ is the number of vertices in the mesh discretization and $C$ is the dimension of signal features. By leveraging the expressive power of CNFs, our approach captures the intrinsic geometry and structure of manifold data while enabling efficient inference and sampling. Our framework can handle complex geometries and severe masking levels across different manifolds, as demonstrated in our experiments with noisy inpainting and imputation tasks (see Section \ref{sec:exps}).

\section{Experiments}
\label{sec:exps}
We validate our approach through various experiments, including (i) Denoising, Super Resolution (SR), and Deblurring with CelebA face image data \citep{liu2015faceattributes}; (ii) Inpainting on Stanford Bunny Manifold with MNIST data; and (iii) Imputation on Sphere with ERA5~\citep{hersbach2020era5} temperature data. (i) Denoising, SR, and Deblurring are performed on the Euclidean in the image domain. In contrast, noisy (ii) inpainting and (iii) imputation are solved directly on the bunny and sphere manifolds. Throughout all the experiments, we use $24$ flow steps and we set the batch size to the $64$. We conduct all the training and optimization experiments on a single RTX3090 GPU instance. Our code is implemented in the JAX framework~\citep{jax2018github}. Necessary details for the experiment settings are given in the Appendix~\ref{detail_tab}~\&~\ref{exp_setting}.

\begin{table}[t]
\centering
\setlength{\tabcolsep}{0.2em}
\resizebox{1.0\textwidth}{!}{
\begin{tabular}{lccc@{\hskip 15pt}ccc@{\hskip 15pt}ccc}
\toprule
{\textbf{Geometry}} & \multicolumn{3}{c}{\textbf{Euclidean}} & \multicolumn{6}{c}{\textbf{Riemannian}} \\
\cmidrule(lr){2-4}
\cmidrule(lr){5-10}
Dataset \& Task & \multicolumn{3}{c}{celebA (denoising)} & \multicolumn{3}{c}{Bunny-MNIST (inpainting)} & \multicolumn{3}{c}{ERA5 (imputation)} \\
\cmidrule(lr){2-4}
\cmidrule(lr){5-7}
\cmidrule(lr){8-10}
Metric & Time[s]$\downarrow$ & PSNR$\uparrow$ & SSIM$\uparrow$ & Time[s]$\downarrow$ & PSNR$\uparrow$ & MSE$\downarrow$ & Time[s]$\downarrow$ & PSNR$\uparrow$ & SSIM$\uparrow$ \\
\midrule
MCG~\citep{chung2022improving} & - & - & - & 19.85 & 26.69 & 0.0024 & 16.16 & 27.52 & 0.871 \\
Noise2Score~\citep{kim2021noise2score} & 0.0172 & 24.36 & 0.871 & - & - & - & - & - & - \\
DPS~\citep{chung2023diffusion} & 16.95 & \textbf{27.93} & \textbf{0.932} & 19.39 & \textbf{28.03} & \textbf{0.0017} & 15.36 & 28.95 & \underline{0.953} \\
\cmidrule(l){1-10}
APS (ours) ($N = 1$) & \textbf{0.0021} & 23.37 & 0.836 & \textbf{0.0021} & 25.97 & 0.0032 & \textbf{0.0012} & \underline{33.17} & 0.883 \\
APS (ours) ($N = 128$) & \underline{0.0035} & \underline{25.82} & \underline{0.901} & \underline{0.0035} & \underline{26.72} & \underline{0.0022} & \underline{0.0018} & \textbf{34.61} & \textbf{0.959} \\
\bottomrule
\end{tabular}
}
\caption{
\add{Quantitative results on our 3 main experiments. \textbf{Best}, \underline{second best}}
}
\label{tab:main_results}
\end{table}

\subsection{Results}
In this section, we provide the general results of each different task described above. We compare APS with the various baselines including, DPS~\citep{chung2023diffusion}, MCG~\citep{chung2022improving}, and Noise2Score~\citep{kim2021noise2score} (see Appendix~\ref{dis}). 
\add{It should be noted that MCG and DPS are identical for denoising, and Noise2Score is only applicable to denoising. In such cases, we do not report the metrics.}
We also demonstrate the comparisons and results against Feng \emph{et al.}~\citep{feng2024variational}. Finally, we experimentally confirm the robustness of APS across different unseen data or datasets (see Appendix~\ref{robust}). For evaluation purposes, we use peak signal-to-noise ratio (PSNR) and structural-similarity-index-measure (SSIM) which are widely used to assess the performance of inverse solvers with the ground truth and reconstructed signals. We further evaluate Fréchet Inception Distance (FID) to showcase the perceptual quality of generated samples. \add{As the proposed method, APS, can sample multiple different posterior samples with a single forward pass, and this process is easily parallelizable, we report two different types for the proposed method. One by taking a single posterior sample ($N = 1$), and another by taking 128 posterior samples and taking the mean ($N = 128$).}

\begin{figure}[!thb]
\centering
\begin{minipage}[m]{0.5\textwidth}
\centering
\resizebox{\textwidth}{!}{
\begin{tabular}{l@{\hskip 15pt}lll@{\hskip 15pt}lll@{\hskip 15pt}l}
\toprule
{} & \multicolumn{3}{c}{\textbf{Gaussian Deblurring}} & \multicolumn{3}{c}{\textbf{Super Resolution x2}} & \\
\cmidrule(lr){2-4}
\cmidrule(lr){5-7}
\cmidrule(lr){8-8}
{\textbf{Method}} & {PSNR $\uparrow$} & {SSIM $\uparrow$}  & {FID $\downarrow$} & {PSNR $\uparrow$} & {SSIM $\uparrow$} & {FID $\downarrow$} & {Time[s] $\downarrow$}\\
\cmidrule(l){1-8}
MCG~\citep{chung2022improving} & 27.38 & 0.928 & 75.63 & 22.64 & 0.859 & 100.3 & 17.27 \\
DPS~\citep{chung2023diffusion} & \textbf{27.40} & \textbf{0.928} & 74.84 & \textbf{24.44} & \textbf{0.865} & 89.92 & 16.95 \\
\cmidrule(l){1-8}
APS (ours) & 26.34 & 0.923 & \textbf{56.11} & 23.81 & 0.860 & \textbf{86.23} & \textbf{0.0035} \\
\bottomrule
\end{tabular}
}
\captionof{table}{
Quantitative results
}
\label{tab:deblur_SRx2_tab}
\end{minipage}%
\hfill
\begin{minipage}[m]{0.5\textwidth}
\centering
\includegraphics[width=\textwidth]{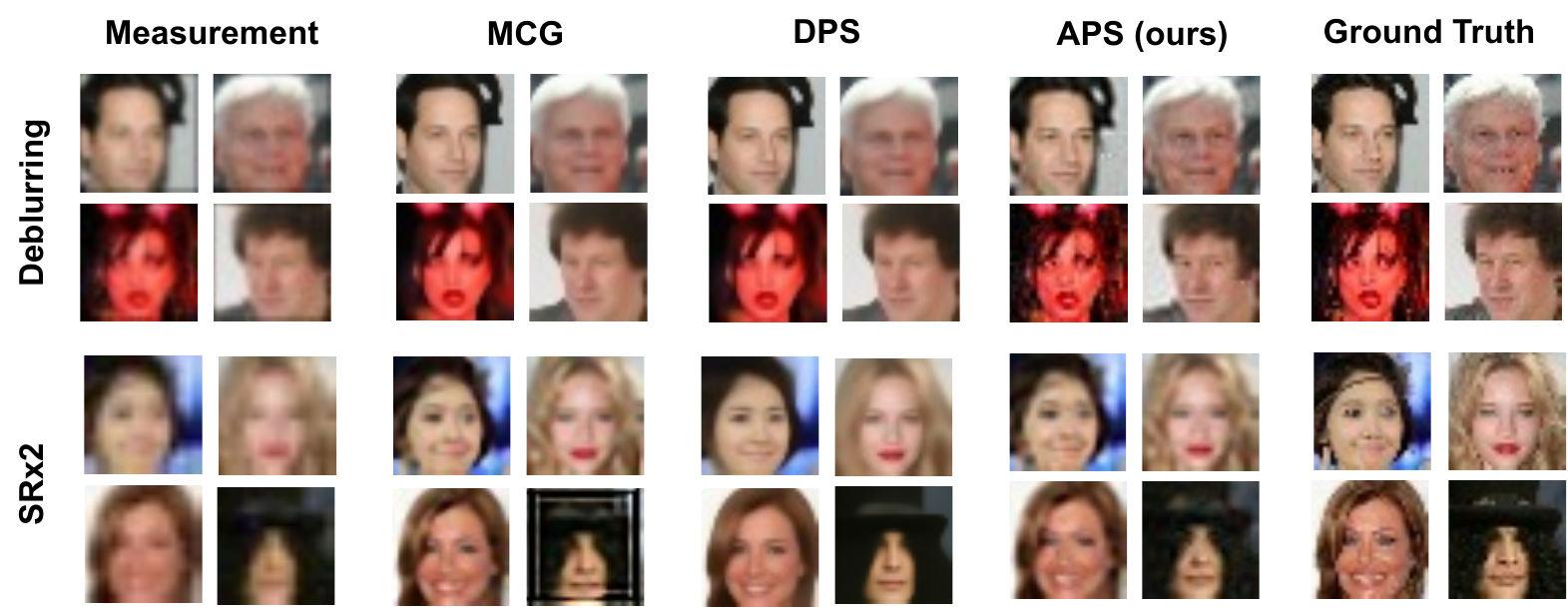}
\caption{Qualitative results}
\label{fig:deblur_SRx2_fig}
\end{minipage}
\vspace{-0.5cm}
\end{figure}

\subsubsection{General Results}
In general, our approach achieves competitive quantitative and qualitative results across different datasets on Euclidean and non-Euclidean geometries. We observe significant time improvements due to the single-step generation ability of our framework. Tab.~\ref{tab:main_results} and Fig.~\ref{fig:main},\ref{fig:baselines} depict competitive quantitative and qualitative results confirming discussions along with instant time generations. \add{It should be noted that the boosted version ($N = 128$) of the proposed method only marginally increases the compute time, as we can sample multiple reconstructions in parallel. To demonstrate that our method can be applied to more general inverse problems, similar to~\citep{chung2023diffusion}, we conducted 2$\times$ Super Resolution and Gaussian Deblurring experiments on the celebA data as shown in Tab.~\ref{tab:deblur_SRx2_tab} \& Fig.~\ref{fig:deblur_SRx2_fig}, where we see that the perceptual quality of the proposed method is {\em better} while being $\sim \times 1000$ faster, and the difference in the distortion metrics are small. Interestingly, while Noise2Score approximates the posterior mean, and the boosted version of the proposed method also approximates the posterior mean by taking the average of the posterior samples, our method outperforms Noise2Score by more than 1 db, showcasing the superiority of the proposed method.}

\subsubsection{Comparing with score prior method}
Compared to the exact score prior~\citep{feng2023score}, surrogate counterpart~\citep{feng2024variational} presents $100$ times faster approach along with competitive or slightly better results in terms of quality. Despite being fast in terms of optimization of NF, \citep{feng2024variational} still requires training the network for a considerable amount of time for every single measurement. We observed that under same conditions, conditional NF does not increase the complexity and training stage takes 0.15 seconds which is 0.14 seconds in case of unconditional version. We further sample a random point from test data of celebA and optimize unconditional NF with the same configurations as ours on this single measurement. NF trained solely on this data reaches 23.75dB in PSNR score, which is almost same as our result of 23.43dB on this measurement. All these confirms that under same conditions, APS can simply achieve best results being also amortized for plenty of measurements.

\section{Discussion}
\label{sec:discussion}

We show a first proof of concept that we can construct a one-step posterior sampler that generalizes across any measurements in an unsupervised fashion (only having access to the measurements $\y$). Notably, APS extends to wide use cases with minimal constraints: 1) the operator $\Ac$ can be arbitrarily complex and non-linear, as in DPS~\citep{chung2023diffusion}, unlike many recent DIS that requires linearity of the operator~\citep{wang2023zeroshot,chung2024decomposed,zhu2023denoising}; 2) training of the sampler can be done without any strict conditions on the measurement, unlike recent unsupervised score training methods that require i.i.d. measurement conditions with the same randomized forward operator~\citep{daras2023ambient,kawar2023gsure}; 3) method can be generalized into different geometries and datasets in a blind manner, unlike recent DIS methods which require to know forward operator during sampling~\citep{chung2023solving, mardani2023variational}. We opted for simplicity in the architecture design of $G_\phi$, and avoided introducing inductive bias of spatial information by taking a vectorized input, potentially explaining the slight background noise in the reconstructions. Further optimization in the choice of network architecture is left as a future direction of study.

\section{Conclusion}
\label{sec:conclusion}

In this work, we propose to use a conditional NF for a VI-based optimization strategy to train a one-step posterior sampler, which implicitly samples from the posterior distribution defined from the pre-trained diffusion prior. We show that APS is highly generalizable, being able to reconstruct samples that are not seen during training, applicable to diverse forward measurements, and types of data, encompassing standard Euclidean geometry as well as data on general Riemannian manifolds. We believe that our work can act as a cornerstone for developing a fast, practical posterior sampler that distills the diffusion prior.

\clearpage

\bibliography{iclr2025_conference}

\begin{thebibliography}{45}
\providecommand{\natexlab}[1]{#1}
\providecommand{\url}[1]{\texttt{#1}}
\expandafter\ifx\csname urlstyle\endcsname\relax
  \providecommand{\doi}[1]{doi: #1}\else
  \providecommand{\doi}{doi: \begingroup \urlstyle{rm}\Url}\fi

\bibitem[Bradbury et~al.(2018)Bradbury, Frostig, Hawkins, Johnson, Leary, Maclaurin, Necula, Paszke, Vander{P}las, Wanderman-{M}ilne, and Zhang]{jax2018github}
James Bradbury, Roy Frostig, Peter Hawkins, Matthew~James Johnson, Chris Leary, Dougal Maclaurin, George Necula, Adam Paszke, Jake Vander{P}las, Skye Wanderman-{M}ilne, and Qiao Zhang.
\newblock {JAX}: composable transformations of {P}ython+{N}um{P}y programs, 2018.
\newblock URL \url{http://github.com/jax-ml/jax}.

\bibitem[Chen et~al.(2018)Chen, Rubanova, Bettencourt, and Duvenaud]{chen2018neural}
Ricky T.~Q. Chen, Yulia Rubanova, Jesse Bettencourt, and David~K Duvenaud.
\newblock Neural ordinary differential equations.
\newblock In \emph{Advances in Neural Information Processing Systems}, volume~31, 2018.

\bibitem[Chung et~al.(2022{\natexlab{a}})Chung, Sim, Ryu, and Ye]{chung2022improving}
Hyungjin Chung, Byeongsu Sim, Dohoon Ryu, and Jong~Chul Ye.
\newblock Improving diffusion models for inverse problems using manifold constraints.
\newblock In Alice~H. Oh, Alekh Agarwal, Danielle Belgrave, and Kyunghyun Cho (eds.), \emph{Advances in Neural Information Processing Systems}, 2022{\natexlab{a}}.
\newblock URL \url{https://openreview.net/forum?id=nJJjv0JDJju}.

\bibitem[Chung et~al.(2022{\natexlab{b}})Chung, Sim, and Ye]{chung2022come}
Hyungjin Chung, Byeongsu Sim, and Jong~Chul Ye.
\newblock {Come-Closer-Diffuse-Faster: Accelerating Conditional Diffusion Models for Inverse Problems through Stochastic Contraction}.
\newblock In \emph{Proceedings of the IEEE/CVF Conference on Computer Vision and Pattern Recognition}, 2022{\natexlab{b}}.

\bibitem[Chung et~al.(2023{\natexlab{a}})Chung, Kim, Mccann, Klasky, and Ye]{chung2023diffusion}
Hyungjin Chung, Jeongsol Kim, Michael~Thompson Mccann, Marc~Louis Klasky, and Jong~Chul Ye.
\newblock Diffusion posterior sampling for general noisy inverse problems.
\newblock In \emph{International Conference on Learning Representations}, 2023{\natexlab{a}}.
\newblock URL \url{https://openreview.net/forum?id=OnD9zGAGT0k}.

\bibitem[Chung et~al.(2023{\natexlab{b}})Chung, Ryu, Mccann, Klasky, and Ye]{chung2023solving}
Hyungjin Chung, Dohoon Ryu, Michael~T Mccann, Marc~L Klasky, and Jong~Chul Ye.
\newblock Solving 3d inverse problems using pre-trained 2d diffusion models.
\newblock \emph{IEEE/CVF Conference on Computer Vision and Pattern Recognition}, 2023{\natexlab{b}}.

\bibitem[Chung et~al.(2024)Chung, Lee, and Ye]{chung2024decomposed}
Hyungjin Chung, Suhyeon Lee, and Jong~Chul Ye.
\newblock Decomposed diffusion sampler for accelerating large-scale inverse problems.
\newblock In \emph{International Conference on Learning Representations}, 2024.

\bibitem[Daras et~al.(2023)Daras, Shah, Dagan, Gollakota, Dimakis, and Klivans]{daras2023ambient}
Giannis Daras, Kulin Shah, Yuval Dagan, Aravind Gollakota, Alexandros~G Dimakis, and Adam Klivans.
\newblock Ambient diffusion: Learning clean distributions from corrupted data.
\newblock \emph{arXiv preprint arXiv:2305.19256}, 2023.

\bibitem[Daras et~al.(2024)Daras, Chung, Lai, Mitsufuji, Ye, Milanfar, Dimakis, and Delbracio]{diffusion_survey}
Giannis Daras, Hyungjin Chung, Chieh-Hsin Lai, Yuki Mitsufuji, Jong~Chul Ye, Peyman Milanfar, Alexandros~G. Dimakis, and Mauricio Delbracio.
\newblock A survey on diffusion models for inverse problems, 2024.
\newblock URL \url{https://arxiv.org/abs/2410.00083}.

\bibitem[Dinh et~al.(2016)Dinh, Sohl-Dickstein, and Bengio]{dinh2016density}
Laurent Dinh, Jascha Sohl-Dickstein, and Samy Bengio.
\newblock Density estimation using real nvp.
\newblock \emph{arXiv preprint arXiv:1605.08803}, 2016.

\bibitem[Dupont et~al.(2021)Dupont, Teh, and Doucet]{dupont2021gasp}
Emilien Dupont, Yee~Whye Teh, and Arnaud Doucet.
\newblock Generative models as distributions of functions.
\newblock \emph{arXiv preprint arXiv:2102.04776}, 2021.

\bibitem[Dupont et~al.(2022{\natexlab{a}})Dupont, Kim, Eslami, Rezende, and Rosenbaum]{dupont2022functa}
Emilien Dupont, Hyunjik Kim, SM~Eslami, Danilo Rezende, and Dan Rosenbaum.
\newblock From data to functa: Your data point is a function and you can treat it like one.
\newblock \emph{arXiv preprint arXiv:2201.12204}, 2022{\natexlab{a}}.

\bibitem[Dupont et~al.(2022{\natexlab{b}})Dupont, Teh, and Doucet]{dupontetal5gasp}
Emilien Dupont, Yee~Whye Teh, and Arnaud Doucet.
\newblock Generative models as distributions of functions.
\newblock In \emph{International Conference on Artificial Intelligence and Statistics}, pp.\  2989--3015. PMLR, 2022{\natexlab{b}}.

\bibitem[Efron(2011)]{efron2011tweedie}
Bradley Efron.
\newblock Tweedie’s formula and selection bias.
\newblock \emph{Journal of the American Statistical Association}, 106\penalty0 (496):\penalty0 1602--1614, 2011.

\bibitem[Feng \& Bouman(2024)Feng and Bouman]{feng2024variational}
Berthy Feng and Katherine Bouman.
\newblock Variational bayesian imaging with an efficient surrogate score-based prior.
\newblock \emph{Transactions on Machine Learning Research}, 2024.
\newblock ISSN 2835-8856.
\newblock URL \url{https://openreview.net/forum?id=db2pFKVcm1}.

\bibitem[Feng et~al.(2023)Feng, Smith, Rubinstein, Chang, Bouman, and Freeman]{feng2023score}
Berthy~T Feng, Jamie Smith, Michael Rubinstein, Huiwen Chang, Katherine~L Bouman, and William~T Freeman.
\newblock Score-based diffusion models as principled priors for inverse imaging.
\newblock In \emph{International Conference on Computer Vision (ICCV)}. IEEE, 2023.

\bibitem[Goodfellow et~al.(2014)Goodfellow, Pouget-Abadie, Mirza, Xu, Warde-Farley, Ozair, Courville, and Bengio]{goodfellow2014generative}
Ian Goodfellow, Jean Pouget-Abadie, Mehdi Mirza, Bing Xu, David Warde-Farley, Sherjil Ozair, Aaron Courville, and Yoshua Bengio.
\newblock Generative adversarial nets.
\newblock \emph{Advances in neural information processing systems}, 27, 2014.

\bibitem[Gu et~al.(2023)Gu, Zhai, Zhang, Liu, and Susskind]{gu2023boot}
Jiatao Gu, Shuangfei Zhai, Yizhe Zhang, Lingjie Liu, and Joshua~M Susskind.
\newblock Boot: Data-free distillation of denoising diffusion models with bootstrapping.
\newblock In \emph{ICML 2023 Workshop on Structured Probabilistic Inference $\{$$\backslash$\&$\}$ Generative Modeling}, 2023.

\bibitem[Hersbach et~al.(2020)Hersbach, Bell, Berrisford, Hirahara, Hor{\'a}nyi, Mu{\~n}oz-Sabater, Nicolas, Peubey, Radu, Schepers, et~al.]{hersbach2020era5}
Hans Hersbach, Bill Bell, Paul Berrisford, Shoji Hirahara, Andr{\'a}s Hor{\'a}nyi, Joaqu{\'\i}n Mu{\~n}oz-Sabater, Julien Nicolas, Carole Peubey, Raluca Radu, Dinand Schepers, et~al.
\newblock The era5 global reanalysis.
\newblock \emph{Quarterly Journal of the Royal Meteorological Society}, 146\penalty0 (730):\penalty0 1999--2049, 2020.

\bibitem[Ho et~al.(2020)Ho, Jain, and Abbeel]{ho2020denoising}
Jonathan Ho, Ajay Jain, and Pieter Abbeel.
\newblock Denoising diffusion probabilistic models.
\newblock \emph{Advances in Neural Information Processing Systems}, 33:\penalty0 6840--6851, 2020.

\bibitem[Hyv{\"a}rinen \& Dayan(2005)Hyv{\"a}rinen and Dayan]{hyvarinen2005estimation}
Aapo Hyv{\"a}rinen and Peter Dayan.
\newblock Estimation of non-normalized statistical models by score matching.
\newblock \emph{Journal of Machine Learning Research}, 6\penalty0 (4), 2005.

\bibitem[Kadkhodaie \& Simoncelli(2021)Kadkhodaie and Simoncelli]{kadkhodaie2020solving}
Zahra Kadkhodaie and Eero Simoncelli.
\newblock Stochastic solutions for linear inverse problems using the prior implicit in a denoiser.
\newblock In \emph{Advances in Neural Information Processing Systems}, volume~34, pp.\  13242--13254. Curran Associates, Inc., 2021.

\bibitem[Karras et~al.(2019)Karras, Laine, and Aila]{karras2019style}
Tero Karras, Samuli Laine, and Timo Aila.
\newblock A style-based generator architecture for generative adversarial networks.
\newblock In \emph{Proceedings of the IEEE/CVF Conference on Computer Vision and Pattern Recognition}, pp.\  4401--4410, 2019.

\bibitem[Kawar et~al.(2022)Kawar, Elad, Ermon, and Song]{kawar2022denoising}
Bahjat Kawar, Michael Elad, Stefano Ermon, and Jiaming Song.
\newblock Denoising diffusion restoration models.
\newblock In Alice~H. Oh, Alekh Agarwal, Danielle Belgrave, and Kyunghyun Cho (eds.), \emph{Advances in Neural Information Processing Systems}, 2022.
\newblock URL \url{https://openreview.net/forum?id=kxXvopt9pWK}.

\bibitem[Kawar et~al.(2023)Kawar, Elata, Michaeli, and Elad]{kawar2023gsure}
Bahjat Kawar, Noam Elata, Tomer Michaeli, and Michael Elad.
\newblock Gsure-based diffusion model training with corrupted data.
\newblock \emph{arXiv preprint arXiv:2305.13128}, 2023.

\bibitem[Kim \& Ye(2021)Kim and Ye]{kim2021noise2score}
Kwanyoung Kim and Jong~Chul Ye.
\newblock {Noise2Score: Tweedie’s Approach to Self-Supervised Image Denoising without Clean Images}.
\newblock \emph{Advances in Neural Information Processing Systems}, 34, 2021.

\bibitem[Liu et~al.(2015)Liu, Luo, Wang, and Tang]{liu2015faceattributes}
Ziwei Liu, Ping Luo, Xiaogang Wang, and Xiaoou Tang.
\newblock Deep learning face attributes in the wild.
\newblock In \emph{Proceedings of International Conference on Computer Vision (ICCV)}, December 2015.

\bibitem[Lu et~al.(2022)Lu, Zhou, Bao, Chen, Li, and Zhu]{lu2022dpm}
Cheng Lu, Yuhao Zhou, Fan Bao, Jianfei Chen, Chongxuan Li, and Jun Zhu.
\newblock {DPM}-solver: A fast {ODE} solver for diffusion probabilistic model sampling in around 10 steps.
\newblock In \emph{Advances in Neural Information Processing Systems}, 2022.
\newblock URL \url{https://openreview.net/forum?id=2uAaGwlP_V}.

\bibitem[Lugmayr et~al.(2020)Lugmayr, Danelljan, Van~Gool, and Timofte]{lugmayr2020srflow}
Andreas Lugmayr, Martin Danelljan, Luc Van~Gool, and Radu Timofte.
\newblock Srflow: Learning the super-resolution space with normalizing flow.
\newblock In \emph{Computer Vision--ECCV 2020: 16th European Conference, Glasgow, UK, August 23--28, 2020, Proceedings, Part V 16}, pp.\  715--732. Springer, 2020.

\bibitem[Luo et~al.(2024)Luo, Hu, Zhang, Sun, Li, and Zhang]{luo2024diff}
Weijian Luo, Tianyang Hu, Shifeng Zhang, Jiacheng Sun, Zhenguo Li, and Zhihua Zhang.
\newblock Diff-instruct: A universal approach for transferring knowledge from pre-trained diffusion models.
\newblock \emph{Advances in Neural Information Processing Systems}, 36, 2024.

\bibitem[Mardani et~al.(2023)Mardani, Song, Kautz, and Vahdat]{mardani2023variational}
Morteza Mardani, Jiaming Song, Jan Kautz, and Arash Vahdat.
\newblock A variational perspective on solving inverse problems with diffusion models.
\newblock \emph{arXiv preprint arXiv:2305.04391}, 2023.

\bibitem[Peng et~al.(2024)Peng, Zheng, Dai, Xiao, Li, Zou, and Xiong]{peng2024improving}
Xinyu Peng, Ziyang Zheng, Wenrui Dai, Nuoqian Xiao, Chenglin Li, Junni Zou, and Hongkai Xiong.
\newblock Improving diffusion models for inverse problems using optimal posterior covariance.
\newblock \emph{arXiv preprint arXiv:2402.02149}, 2024.

\bibitem[Poole et~al.(2023)Poole, Jain, Barron, and Mildenhall]{poole2023dreamfusion}
Ben Poole, Ajay Jain, Jonathan~T. Barron, and Ben Mildenhall.
\newblock Dreamfusion: Text-to-3d using 2d diffusion.
\newblock In \emph{The Eleventh International Conference on Learning Representations}, 2023.
\newblock URL \url{https://openreview.net/forum?id=FjNys5c7VyY}.

\bibitem[Rezende \& Mohamed(2015)Rezende and Mohamed]{rezende2015variational}
Danilo Rezende and Shakir Mohamed.
\newblock Variational inference with normalizing flows.
\newblock In \emph{International conference on machine learning}, pp.\  1530--1538. PMLR, 2015.

\bibitem[Song et~al.(2021{\natexlab{a}})Song, Meng, and Ermon]{song2020denoising}
Jiaming Song, Chenlin Meng, and Stefano Ermon.
\newblock Denoising diffusion implicit models.
\newblock In \emph{9th International Conference on Learning Representations, {ICLR}}, 2021{\natexlab{a}}.

\bibitem[Song et~al.(2023{\natexlab{a}})Song, Vahdat, Mardani, and Kautz]{song2023pseudoinverseguided}
Jiaming Song, Arash Vahdat, Morteza Mardani, and Jan Kautz.
\newblock Pseudoinverse-guided diffusion models for inverse problems.
\newblock In \emph{International Conference on Learning Representations}, 2023{\natexlab{a}}.
\newblock URL \url{https://openreview.net/forum?id=9_gsMA8MRKQ}.

\bibitem[Song \& Ermon(2019)Song and Ermon]{song2019generative}
Yang Song and Stefano Ermon.
\newblock Generative modeling by estimating gradients of the data distribution.
\newblock In \emph{Advances in Neural Information Processing Systems}, volume~32, 2019.

\bibitem[Song et~al.(2021{\natexlab{b}})Song, Durkan, Murray, and Ermon]{song2021maximum}
Yang Song, Conor Durkan, Iain Murray, and Stefano Ermon.
\newblock Maximum likelihood training of score-based diffusion models.
\newblock \emph{Advances in Neural Information Processing Systems}, 34, 2021{\natexlab{b}}.

\bibitem[Song et~al.(2021{\natexlab{c}})Song, Sohl{-}Dickstein, Kingma, Kumar, Ermon, and Poole]{song2020score}
Yang Song, Jascha Sohl{-}Dickstein, Diederik~P. Kingma, Abhishek Kumar, Stefano Ermon, and Ben Poole.
\newblock Score-based generative modeling through stochastic differential equations.
\newblock In \emph{9th International Conference on Learning Representations, {ICLR}}, 2021{\natexlab{c}}.

\bibitem[Song et~al.(2023{\natexlab{b}})Song, Dhariwal, Chen, and Sutskever]{song2023consistency}
Yang Song, Prafulla Dhariwal, Mark Chen, and Ilya Sutskever.
\newblock Consistency models.
\newblock \emph{arXiv preprint arXiv:2303.01469}, 2023{\natexlab{b}}.

\bibitem[Sun \& Bouman(2021)Sun and Bouman]{sun2021deep}
He~Sun and Katherine~L Bouman.
\newblock Deep probabilistic imaging: Uncertainty quantification and multi-modal solution characterization for computational imaging.
\newblock In \emph{Proceedings of the AAAI Conference on Artificial Intelligence}, volume~35, pp.\  2628--2637, 2021.

\bibitem[Turk \& Levoy(1994)Turk and Levoy]{turk1994zippered}
Greg Turk and Marc Levoy.
\newblock Zippered polygon meshes from range images.
\newblock In \emph{Proceedings of the 21st annual conference on Computer graphics and interactive techniques}, pp.\  311--318, 1994.

\bibitem[Wang et~al.(2023)Wang, Yu, and Zhang]{wang2023zeroshot}
Yinhuai Wang, Jiwen Yu, and Jian Zhang.
\newblock Zero-shot image restoration using denoising diffusion null-space model.
\newblock In \emph{The Eleventh International Conference on Learning Representations}, 2023.
\newblock URL \url{https://openreview.net/forum?id=mRieQgMtNTQ}.

\bibitem[Wang et~al.(2024)Wang, Lu, Wang, Bao, Li, Su, and Zhu]{wang2024prolificdreamer}
Zhengyi Wang, Cheng Lu, Yikai Wang, Fan Bao, Chongxuan Li, Hang Su, and Jun Zhu.
\newblock Prolificdreamer: High-fidelity and diverse text-to-3d generation with variational score distillation.
\newblock \emph{Advances in Neural Information Processing Systems}, 36, 2024.

\bibitem[Zhu et~al.(2023)Zhu, Zhang, Liang, Cao, Wen, Timofte, and Van~Gool]{zhu2023denoising}
Yuanzhi Zhu, Kai Zhang, Jingyun Liang, Jiezhang Cao, Bihan Wen, Radu Timofte, and Luc Van~Gool.
\newblock Denoising diffusion models for plug-and-play image restoration.
\newblock In \emph{Proceedings of the IEEE/CVF Conference on Computer Vision and Pattern Recognition}, pp.\  1219--1229, 2023.

\end{thebibliography}
\bibliographystyle{iclr2025_conference}

\clearpage
\appendix
\section{Appendix}
\label{sec:appendix}

\subsection{Reproducibility and Details of Parameters}
\label{detail_tab}

We further provide all the necessary details to replicate the results with our proposed method. Tab.~\ref{tab:params} demonstrates the details of tasks and datasets along with the parameter choices for both prior score network training and Conditional NF optimization. Note that, we have validated our approach through $3$ different inverse problems on the image dataset (celebA), where noise level was set to 0.1 for denoising and 0.01 for Super-Resolution and Gaussian Deblurring. In addition to this, we also submit all the codes as a supplementary material to reproduce the results shown throughout the paper.

\begin{table}[!hbt]
\vskip 0.15in
\begin{center}
\begin{small}
\begin{tabular}{lcccr}
\toprule
Parameter & CelebA & Bunny  & Sphere \\
\midrule
resolution (\#vertices) & 32 $\times$ 32 & 1889 & 4140 \\ 
distribution on manifold & - & MNIST & ERA5 \\
task & varying & Inpainting & Imputation \\
mask level & - & 30\% & 60\% \\
noise level & varying & 0.1 & 0.05 \\
\#channels & 3 & 1 & 1 \\ 
normalized range & [0, 1] & [0, 1] & [0, 1] \\
\midrule
(S) \#train data & 162,770 & 60,000 & 8,510 \\
(S) batch size & 128 & 64 & 64 \\
(S) learning rate & 2e-4 & 2e-4 & 2e-4\\
(S) \#training iters & 1M & 500k & 360k \\
\midrule
(C) \#test data & 19,962 & 10,000 & 2,420 \\
(C) batch size & 64 & 64 & 64\\
(C) learning rate & 1e-5 & 1e-5 & 1e-5\\
(C) \#optimization iters & 1M & 1.5M & 315k \\
\bottomrule
\\
\end{tabular}
\caption{Different configurations of hyperparameter choices for varying datasets and manifolds learned by APS. (S) and (C) denotes the parameter choices for score network and CNF optimization, respectively.}
\label{tab:params}
\end{small}
\end{center}
\vskip -0.1in
\end{table}

\subsection{Experimental Settings}
\label{exp_setting}
\noindent\textbf{Inverse Problems on CelebA.} We follow the usual formulation and adapt $32\times 32$ resolution of facial images. Data is normalized into [0, 1] range and measurement is acquired by the appropriate choices of forward operator depending on the task (See Appendix~\ref{sec:appendix} for details). $G_\phi$ is optimized over the $19,962$ \textbf{test} images by using the forward operator and prior from diffusion models. We optimize APS for $1$M iterations for all different tasks (convergency was observed earlier but continued for potential refinement).

\noindent\textbf{Inpainting on Bunny MNIST.} In order to demonstrate the geometric awareness of our model, we conduct experiment on Stanford Bunny Manifold. We choose the mesh resolution of $1889$ vertices and then project the [0, 1] normalized MNIST digits onto the bunny manifold \citep{turk1994zippered}. In order to ensure the dimensionality compatibility for the models, we use $1888$ vertices and zero mask the last vertex throughout the experiments. We obtain the measurement by occluding $30\%$ of vertices randomly and adding some Gaussian noise, i.e. $\Ac$ is the random masking operator and $\sigma_y=0.1$ in (\ref{eq:ip}). APS is optimized on the test chunk of $10,000$ digit examples for $1.5$M iterations.

\noindent\textbf{Imputation on ERA5.} To show the essence and practical importance of our pipeline, we further conduct experiments on ERA5 temperature dataset. Even though data is available in a rectangular format, due to the spherical shape of Earth, it inherits some geometric information. We use $4^\circ$ resolution dataset with $4140$ vertices borrowed from \citep{dupontetal5gasp} with only temperature channel as it is quite popular to analyze in the domain of generative AI~\citep{dupont2021gasp, dupont2022functa}. Again, due to the dimensionality, we add $20$ more vertices with a signal value of zero, and the data is [0, 1] normalized. In contrast to Bunny MNIST, we use more severe occlusion of $60\%$ random masking with additional Gaussian noise of $\sigma_y=0.05$. We perform the optimization of APS on the test part of the dataset with $2420$ examples for $315$k iterations.

\noindent\textbf{Score Networks.} For all the diffusion priors, VPSDE formulation has been adapted. In the case of image domain CelebA, we borrow the same score checkpoint used in the~\citep{feng2023score, feng2024variational}, which uses NCSN++ \citep{song2021maximum} architecture and has been trained for $1$M iterations. For Bunny MNIST, we adapt the $1$D formulation of DDPM \citep{ho2020denoising} and train the score network for $500$k, at which the convergence was clearly observed through the generated samples. In the case of spherical weather data, we followed the same strategy as Bunny MNIST but achieved convergence of score network earlier at $360$k iterations.

\subsection{Qualitative comparisons with baselines}
\begin{figure}[!htb]
\centering
\includegraphics[width=\linewidth]{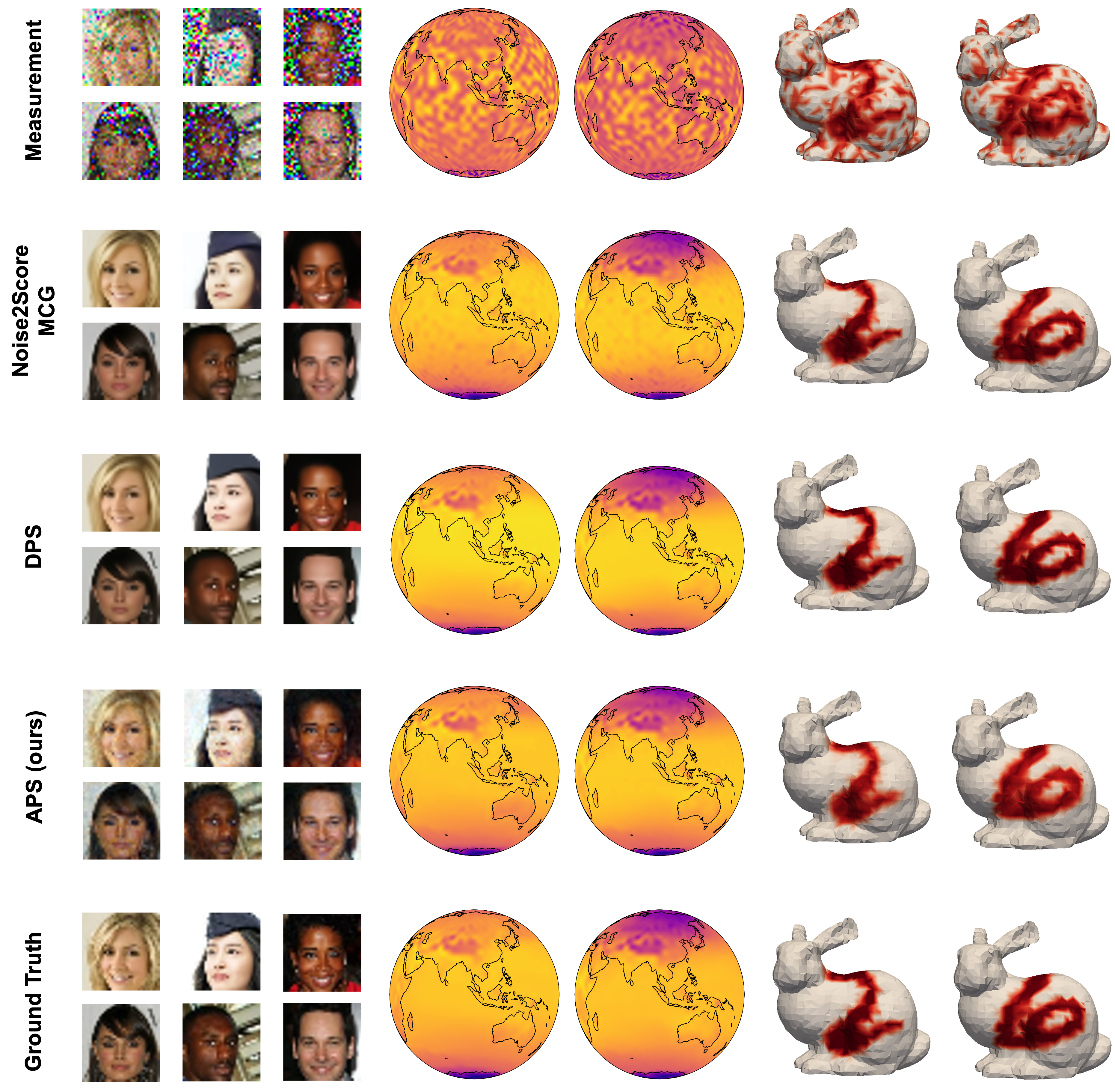}
\caption{Comparisons with different baselines for CelebA denoising, ERA5 imputation, and Bunny MNIST inpainting tasks. Note that second row shows the results of Noise2Score for the CelebA denoising task and MCG for inpainting and imputation of manifold data.}
\label{fig:baselines}
\end{figure}

\subsection{Further Experimental Results}
\label{more_exps}

\subsubsection{Comparing with DIS and Noise2Score}
\label{dis}
\begin{figure}[!thb]
\centering
\includegraphics[width=\linewidth]{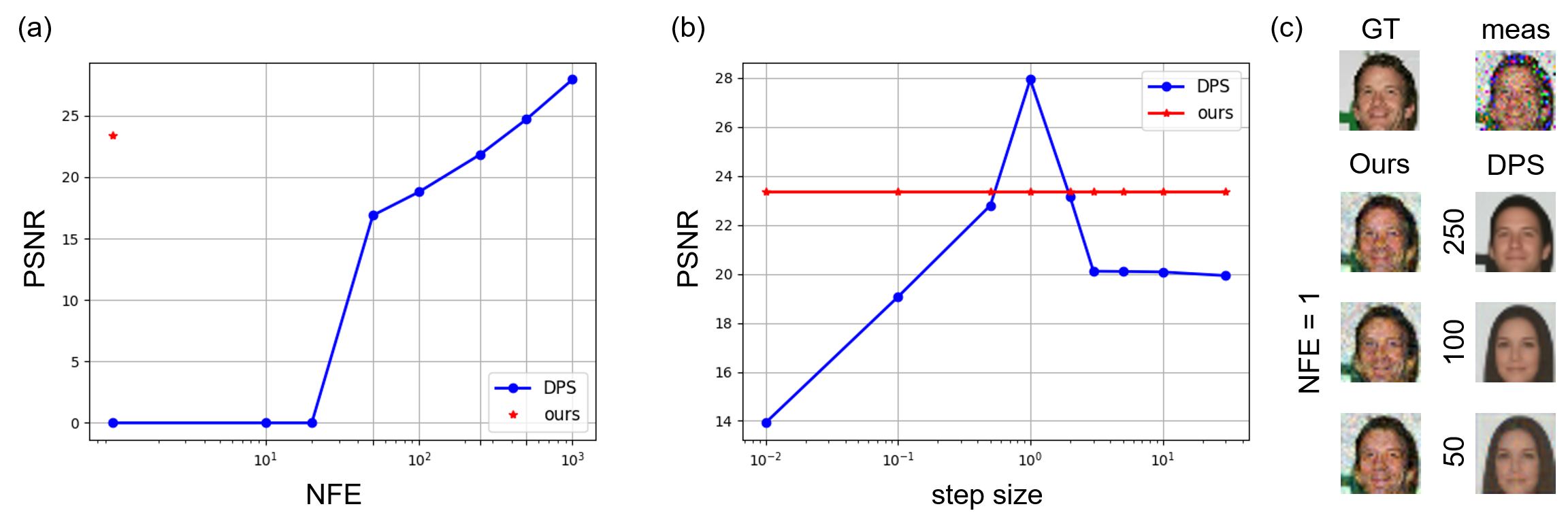}
\caption{Comparison of our method against DPS~\citep{chung2023diffusion} on celebA denoising. (a) NFE vs. PSNR plot, (b) step size (used in DPS only) vs. PSNR plot, (c) representative results by varying the NFE.}
\label{fig:comparison_dps_stability}
\end{figure}

Both DPS~\citep{chung2023diffusion} and MCG~\citep{chung2022improving} leverage the pre-trained diffusion model to sample from the posterior distribution. However, these methods require thousands of NFEs to achieve stable performance. The required time for DPS and MCG is reported in Tab.~\ref{tab:main_results}, \ref{tab:deblur_SRx2_tab}. When decreasing the NFE as shown in Fig.~\ref{fig:comparison_dps_stability} (a), PSNR heavily degrades and eventually diverges when we take an NFE value of less than 30. APS achieves competitive performance even with a single NFE.
Moreover, it is shown in Fig.~\ref{fig:comparison_dps_stability} (b) that even slightly incorrectly choosing the step size parameter leads to a large degradation in performance, whereas our method is free from such cumbersome hyperparameter tuning. Finally, it is shown in Fig.~\ref{fig:comparison_dps_stability} (c) that DPS collapses to the mean of the prior distribution, altering the content of the measurement heavily when we take a smaller number of NFEs.

It is worth mentioning that Noise2Score~\citep{kim2021noise2score} is applicable for one-step denoising of the measurements by leveraging the Tweedie's formula. However, as discussed in Sec.~\ref{sec:discussion}, APS is generally applicable to a wide class of inverse problems, whereas the applicability of Noise2Score is limited.

\subsubsection{Generalizability across datasets and Blind Inverse Problems}
\label{robust}

\begin{figure}[!htb]
\centering
\includegraphics[width=\linewidth]{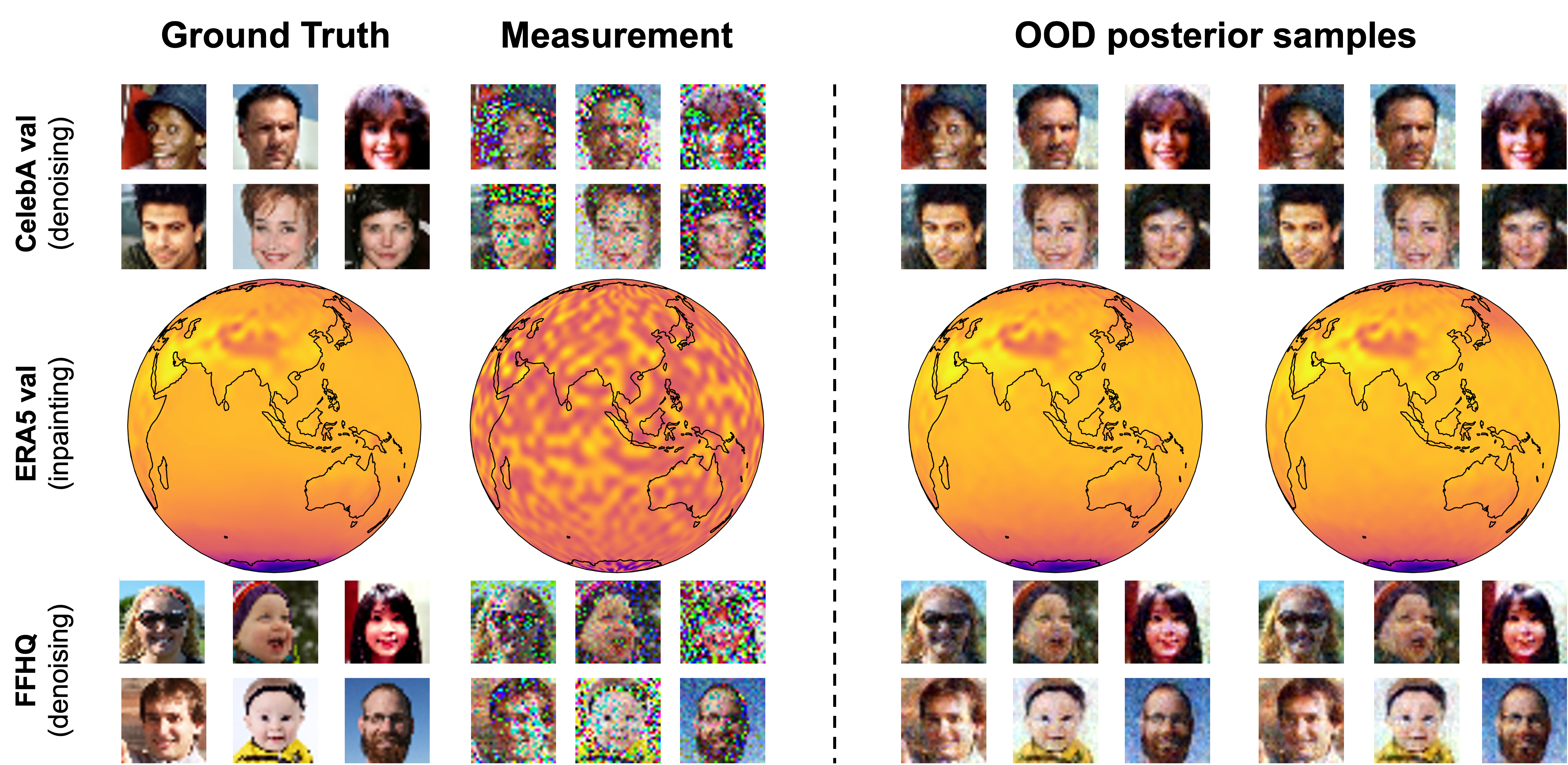}
\caption{Figure demonstrating the visual results of robustness and generalization ability of APS. First and second rows show the results on unseen validation data, and third row depicts generalization to another dataset. Corresponding quantitative analysis can be found in Tab.~\ref{tab:robust_data}.}
\label{fig:robust}
\end{figure}

\begin{table}[!htb]
\centering
\setlength{\tabcolsep}{0.2em}
\resizebox{0.8\textwidth}{!}{
\begin{tabular}{lccc@{\hskip 15pt}ccc@{\hskip 15pt}ccc}
\toprule
Dataset \& Task & {celebA val (denoising)} & {ERA5 val (imputation)} & {FFHQ (denoising)} \\
\cmidrule(lr){2-2}
\cmidrule(lr){3-3}
\cmidrule(lr){4-4}
\midrule
PSNR$\uparrow$ & 23.26 & 33.12 & 21.92 \\
SSIM$\uparrow$ & 0.831 & 0.882 & 0.822 \\
\bottomrule
\end{tabular}
}
\caption{
APS is robust against unseen data samples and even generalizable accross different datasets once it is optimized. For the first $2$ columns, we use validation part of datasets and feed-forward our CNF on this totally unseen data. For the third column, we even show that pre-optimized APS can be leveraged to restore back the noised data samples from across various datasets. All the quantitative results align with their counterparts in Tab.~\ref{tab:main_results} that confirms robustness and generalization ability of our pipeline which was not possible before.}
\label{tab:robust_data}
\vspace{-0.3cm}
\end{table}

We further observe that our optimized framework can be used on unseen data as well. Tab.~\ref{tab:robust_data} and Fig.~\ref{fig:robust} depicts that we achieve similar quantitative and qualitative reconstruction results when we sample from unseen celebA or ERA5 validation datasets. Note that score network is trained on train signals and CNF optimization has been conducted on test signals, i.e. validation is totally hidden to both teacher and student models. More strongly, our approach can also be leveraged on various datasets as pre-optimized inverse solver. To this end, we use our celebA optimized CNF model to perform Denoising task on the FFHQ~\citep{karras2019style} dataset. Same Table and Figure show that our model can remove the noise artifacts with a similar performance as it does on original data, confirming the generalizibility feature.

\begin{figure}[t]
  \centering
  \centerline{{\includegraphics[width=0.8\linewidth]{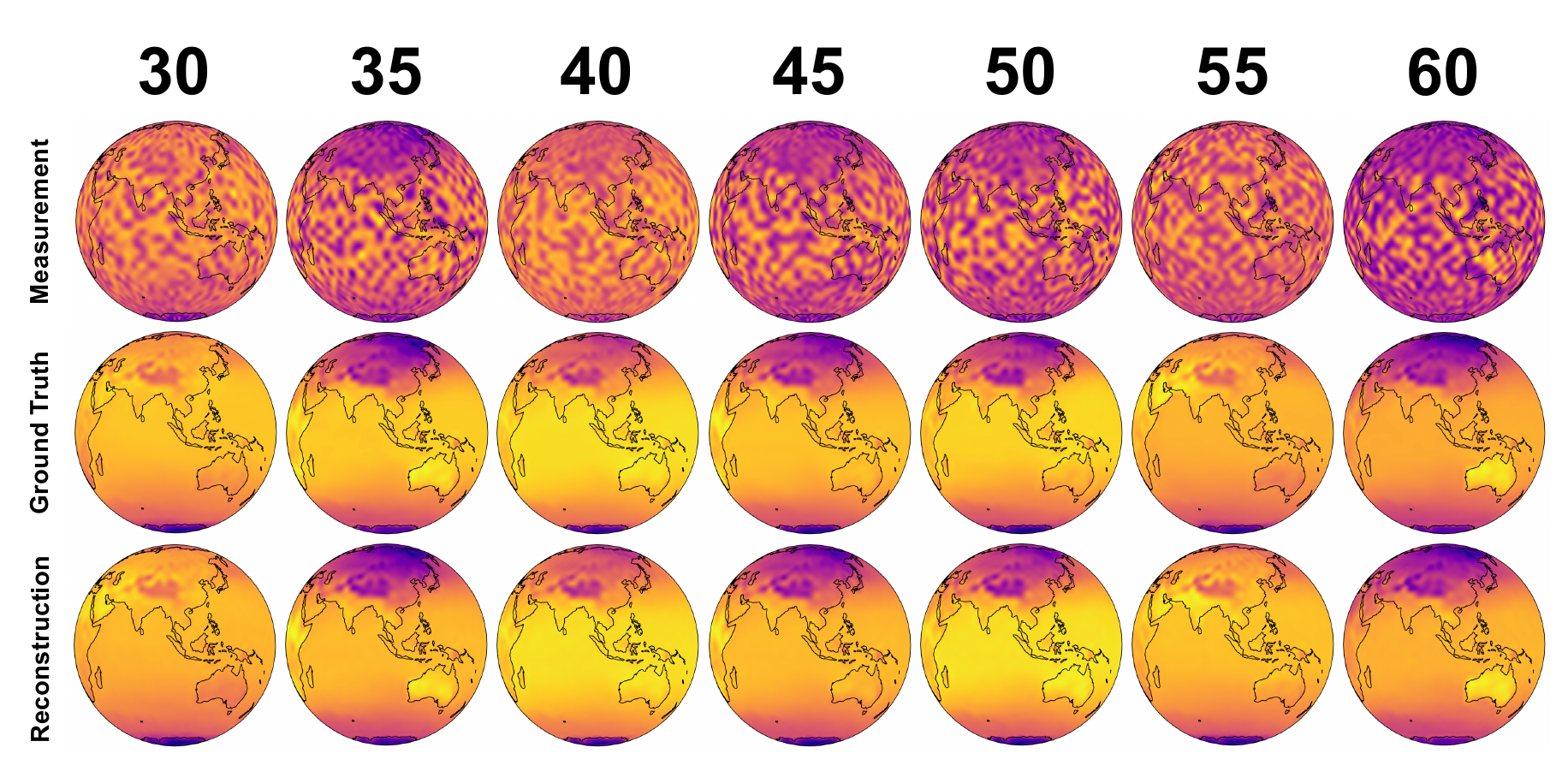}}}
  \caption{Blind inverse problem solving with varying imputation levels.}
  \label{fig:varying_masks_blind}
  \vspace{-0.2cm}
\end{figure}

We further observed that APS can work in the absence of forward operator. In other words, we can perform blind inverse problems through our amortized posterior sampling. We used various imputation levels between 30\% to 60\% for ERA5 dataset, and conducted experiments with random choice of imputation in a blind manner. As a result, Fig.~\ref{fig:varying_masks_blind} shows that results as good as the original inverse solver with the known forward operation (at least $33$ PSNR across all different blind imputation levels).

\end{document}